\documentclass[a4paper,conference]{IEEEtran}
\IEEEoverridecommandlockouts
\usepackage{hyperref}
\usepackage{times,amsmath,amssymb,amsfonts,amsthm}
\usepackage{algorithmic}
\usepackage{graphicx}
\usepackage{xcolor}
\def\BibTeX{{\rm B\kern-.05em{\sc i\kern-.025em b}\kern-.08em
    T\kern-.1667em\lower.7ex\hbox{E}\kern-.125emX}}

\newcommand{\ii}{\boldsymbol{i}}
\newcommand{\step}{\mathtt{step}}
\newcommand{\sign}{\mathtt{sign}}
\newcommand{\ceil}{\mathtt{ceil}}
\newcommand{\csign}{\mathtt{csign}}
\newcommand{\splitsign}{\mathtt{split-sign}}
\newcommand{\coceil}{\mathtt{coceil}}
\newcommand{\cosign}{\mathtt{CoSign}}

\newtheorem{Theorem}{Theorem}

\theoremstyle{definition}

\begin{document}

\title{Novel Complex-Valued Hopfield Neural Networks with Phase and Magnitude Quantization}


\author{
\IEEEauthorblockN{Garimella Ramamurthy\IEEEauthorrefmark{1}, Marcos Eduardo Valle\IEEEauthorrefmark{2}, and Tata Jagannadha Swamy\IEEEauthorrefmark{3}}
\IEEEauthorblockA{\IEEEauthorrefmark{1}Department of Computer Science and Engineering \\ Mahindra University, Hyderabad, India. \\ Email: rama.murthy@mahindrauniversity.edu.in}
\IEEEauthorblockA{\IEEEauthorrefmark{2}Department of Applied Mathematics \\ Universidade Estadual de Campinas (UNICAMP),  Campinas, Brazil. \\ Email: valle@ime.unicamp.br}
\IEEEauthorblockA{\IEEEauthorrefmark{3}Department of Electronics and Communication Engineering \\ Gokaraju Rangraju Institute of Engineering and Technology, Hyderabad, India. \\ Email: jagan.tata@griet.ac.in}
}

\maketitle

\begin{abstract}
This research paper introduces two novel complex-valued Hopfield neural networks (CvHNNs) that incorporate phase and magnitude quantization. The first CvHNN employs a ceiling-type activation function that operates on the rectangular coordinate representation of the complex net contribution. The second CvHNN similarly incorporates phase and magnitude quantization but utilizes a ceiling-type activation function based on the polar coordinate representation of the complex net contribution. The proposed CvHNNs, with their phase and magnitude quantization, significantly increase the number of states compared to existing models in the literature, thereby expanding the range of potential applications for CvHNNs.  
\end{abstract}


\section{Introduction}
Real-valued neural networks are primarily based on the McCulloch-Pitts model of neurons \cite{mcculloch43, haykin09}. This model represents inputs, outputs, synaptic weights, and biases as real-valued numbers. In the early 1970s, Aizenberg and his collaborators expanded this traditional model by incorporating complex-valued inputs, outputs, synaptic weights, and biases \cite{aizenberg71, Aizenberg2011Complex-ValuedNeurons}. This extension led to the development of complex-valued neural networks, which have been extensively studied by researchers in various contexts and remain an active area of research today \cite{Hirose2009Complex-valuedOrigins,Hirose2012Complex-valuedNetworks,Trabelsi17complex,Voigtlaender2023TheNetworks,Scardapane2020Complex-ValuedFunctions,Lee2022Complex-ValuedSurvey,Wolff2025CVKAN:Networks}. 

Motivated by the challenges of emulating biological memory, Hopfield proposed a recurrent neural network capable of implementing associative memories \cite{Hopfield82,mceliece87}. The Hopfield neural network employs the McCulloch-Pitts model of the neuron and utilizes Hebbian learning to store patterns in its associative memory \cite{hassoun95,hassoun97}. Additionally, Hopfield explored the dynamics of his proposed neural network using concepts from mechanical statistics \cite{Hopfield82}. The stability of the Hopfield neural network is a crucial feature for effectively implementing associative memory. In this regard, Goles and Fogelman provided an insightful approach to proving the convergence theorem associated with the Hopfield neural network \cite{Goles-Chacc1985DecreasingNetworks}. The stable characteristics of the Hopfield neural network play a crucial role in its successful applications, which include solving optimization problems and performing image segmentation tasks \cite{shen97,smith98,pajares06,pajares10,Tolmachev2020NewMinimisation}. Despite being proposed in the early 1980s, Hopfield neural networks remain a topic of research today, partly due to their connection with the transformer architecture \cite{Krotov2020LargeLearning,Ramsauer2020HopfieldNeed,Karakida2024HierarchicalBreaking}.

Complex-valued neural networks (CvNNs) are powerful models for addressing problems involving complex-valued signals and functions of complex variables. One key advantage lies in their ability to effectively process and preserve phase information \cite{Aizenberg2011Complex-ValuedNeurons}. Moreover, CvNNs offer enhanced functionality, plasticity, and flexibility compared to traditional real-valued models, enabling faster learning and better generalization. Leveraging these strengths, this paper introduces novel complex-valued Hopfield neural network (CvHNN) models.

The development of CvHNNs began in the late 1980s with the pioneering work of Noest \cite{noest88a,noest88b,noest88c}. Later, Jankowski et al. proposed a multi-valued associative memory utilizing CvHNNs based on the complex-signum activation function \cite{Jankowski1996Complex-valuedMemory}. In this framework, the states of the CvHNN are represented by unit complex numbers, and the complex-signum function normalizes (sets the magnitude to unit) and quantizes the phase of the net contribution. Building on this foundation, Muezzinoglu et al. introduced a design method for CvHNNs using matrix inequalities, with applications in image processing \cite{Muezzinoglu2003AMemory}. Subsequent enhancements include improving storage capacity through the projection rule \cite{lee06} and increasing noise tolerance by replacing the complex-signum function with soft variants, such as the complex-sigmoid function, or adopting the multistate bifurcating neuron model \cite{tanaka09}. Recent advances focus on accelerating training and further improving noise tolerance \cite{Hashimoto2025EnhancingParallelization,Kobayashi2025Complex-valuedConnections}. Moreover, theoretical and practical progress in CvHNNs has also inspired extensions to high-dimensional hypercomplex algebras, such as quaternions \cite{Isokawa2013QuaternionicMemories}, split-quaternions \cite{Kobayashi2020SplitNetworks}, the Klein 4-group \cite{Kobayashi2020HopfieldFour-group}, Clifford algebras \cite{vallejo08}, and Cayley-Dickson algebras \cite{castro20nn}, and vector-valued Hopfield neural networks \cite{Garimella2023Vector-ValuedMemories}.

Despite advancements in storage capacity and noise tolerance, existing CvHNN models primarily rely on phase quantization. In contrast, Garimella and Praveen introduced CvHNN models that utilize magnitude quantization \cite{Murthy2004Complex-valuedHypercube}. However, magnitude quantization introduces unique challenges in analyzing the dynamics of CvHNNs. These challenges, including convergence properties, were addressed in \cite{Murthy2004Complex-valuedHypercube,RamaMurthy2004InfiniteHypercube}.

%
%

This paper advances the field by proposing novel CvHNN models integrating phase and magnitude quantization. We also investigate activation functions in both Cartesian and polar representations of the net contribution. Alongside introducing these innovative CvHNN models, we present key insights to validate their dynamics. This research paper is organized as follows: Section \ref{sec:CvHNNs} provides a brief review of CvHNNs from the literature, focusing on their state space and activation functions. Section \ref{sec:magnitude_phase} presents two novel CvHNNs based on the magnitude and phase quantization. 
Section \ref{sec:experiments} presents computational experiments that offer insights into the proposed network dynamics. The paper finishes with concluding remarks in Section \ref{sec:concluding}.


\section{Complex-Valued Hopfield Neural Networks} \label{sec:CvHNNs}

This section provides a brief overview of complex-valued Hopfield neural networks (CvHNNs). Specifically, we review the fundamental equations that describe the dynamics of CvHNNs. We also discuss two different activation functions used in these models: the complex signum activation function \cite{aizenberg92,Jankowski1996Complex-valuedMemory} and the real-imaginary-type sign activation function \cite{Murthy2004Complex-valuedHypercube,Hirose2012Complex-valuedNetworks}.

As the name implies, the synaptic weights, thresholds, and neuron states in a CvHNN are all complex numbers. Throughout the paper, a complex number is denoted by $z = a + b\ii \in \mathbb{C}$, where $\ii$ represents the imaginary unit with $\ii^2 = -1$. In this expression, $a$ and $b$ denote the real and the imaginary part of $z$, respectively. The conjugate of $z$ is the complex number $\bar{z}=a-b\ii$. Besides the Cartesian representation, a complex number $z$ can be represented alternatively using the polar form as $z = r e^{\theta \ii}$, where $r \geq 0$ and $\theta \in [0,2\pi)$ are the magnitude and phase of $z$, respectively. Recall that the magnitude and the phases are related to the real and imaginary parts through the equations $r = \sqrt{a^2+b^2}$ and $\theta = \arctan(b/a)$.

\subsection{Dynamics of CvHNNs}

Consider a network with \( N \) neurons. The states of the neurons belong to a discrete subset $\mathcal{S}$ of the complex numbers $\mathbb{C}$, corresponding to the image set of the neurons' activation function. The state of the network at time step \( t \) is represented by a vector 
\begin{equation}
\label{eq:state}
\boldsymbol{S}(t) = \big(S_1(t),S_2(t),\ldots,S_N(t) \big) \subset \mathcal{S}^N,
\end{equation}
where $S_i$ corresponds to the state of the $i$th neuron. We arrange the synaptic weights \( W_{ij} \) into a matrix \( \boldsymbol{W} \in \mathbb{C}^{N \times N} \) and the neuron thresholds \( T_i \) into a vector \( \boldsymbol{T} \in \mathbb{C}^{N} \). The update rule for complex neurons is similar to that of real neurons. Considering an activation function $\psi:\mathcal{D} \to \mathcal{S}$, where $\mathcal{D} \in \mathbb{C}$ is the function domain, the $i$th neuron is updated as follows
\begin{equation}
    \label{eq:update}
    S_i(t+1) = \psi \left(\sum_{j=1}^N W_{ij}S_j(t) - T_i \right),
\end{equation}
if the argument belongs to $\mathcal{D}$. Otherwise, the $i$th neuron remains in its current state \cite{Castro2018SomeNetworks,castro20nn}. The update rule specified by \eqref{eq:update} can be implemented in one of two ways:
\begin{itemize}
    \item \textbf{Serial mode:} In this approach, a single neuron is updated at each step. In other words, the state of a single neuron is updated using \eqref{eq:update}, for some $i \in \{1,\ldots,N\}$.
    \item \textbf{(Fully) Parallel mode:} In this method, all neurons are updated simultaneously based on the current state of the network. In mathematical terms, the neuron's states are updated using $\eqref{eq:update}$ for all $i=1,\ldots,N$.
\end{itemize}
The parallel mode, in which all neurons are updated simultaneously, is also called the \textbf{synchronous} update. Additionally, the serial mode is known as the 	\textbf{asynchronous} update.

A CvHNN is said to stabilize at a \textbf{cycle of length $L$} if there is $t_0 \geq 0$ such that $S(t+L)=S(t)$ for all $t \geq t_0$, meaning the network returns to the same state after $L$ steps. Furthermore, we say the network converges to a \textbf{stable state} if it stabilizes at a cycle with a length $L=1$. In simpler terms, the network converges to a stable state when all neurons maintain their state, which implies that there exists a time $t_0 \geq 0$ such that $S(t+1)=S(t)$ for every $t \geq t_0$. 

The update mode plays a key role in the dynamics of Hopfield neural networks \cite{Bruck1988ANetworks,Castro2018SomeNetworks,Hopfield82}. Indeed, a network operating in serial mode may settle at a stable state, but the same network  may not stabilize in parallel update mode. One usually presents an energy function to analyze the dynamics of a recurrent neural network. Briefly, an energy function is a bounded real-valued mapping on the set of all network states. Moreover, it must decrease when evaluated on consecutive but different network states. The energy function depends on several aspects of the network, including the operation mode \cite{Hopfield82,Bruck1988ANetworks,Goles-Chacc1985DecreasingNetworks,Murthy2004Complex-valuedHypercube}.

\subsection{Complex Signum Function}

\begin{figure}
    \centering
    \includegraphics[width=0.9\linewidth]{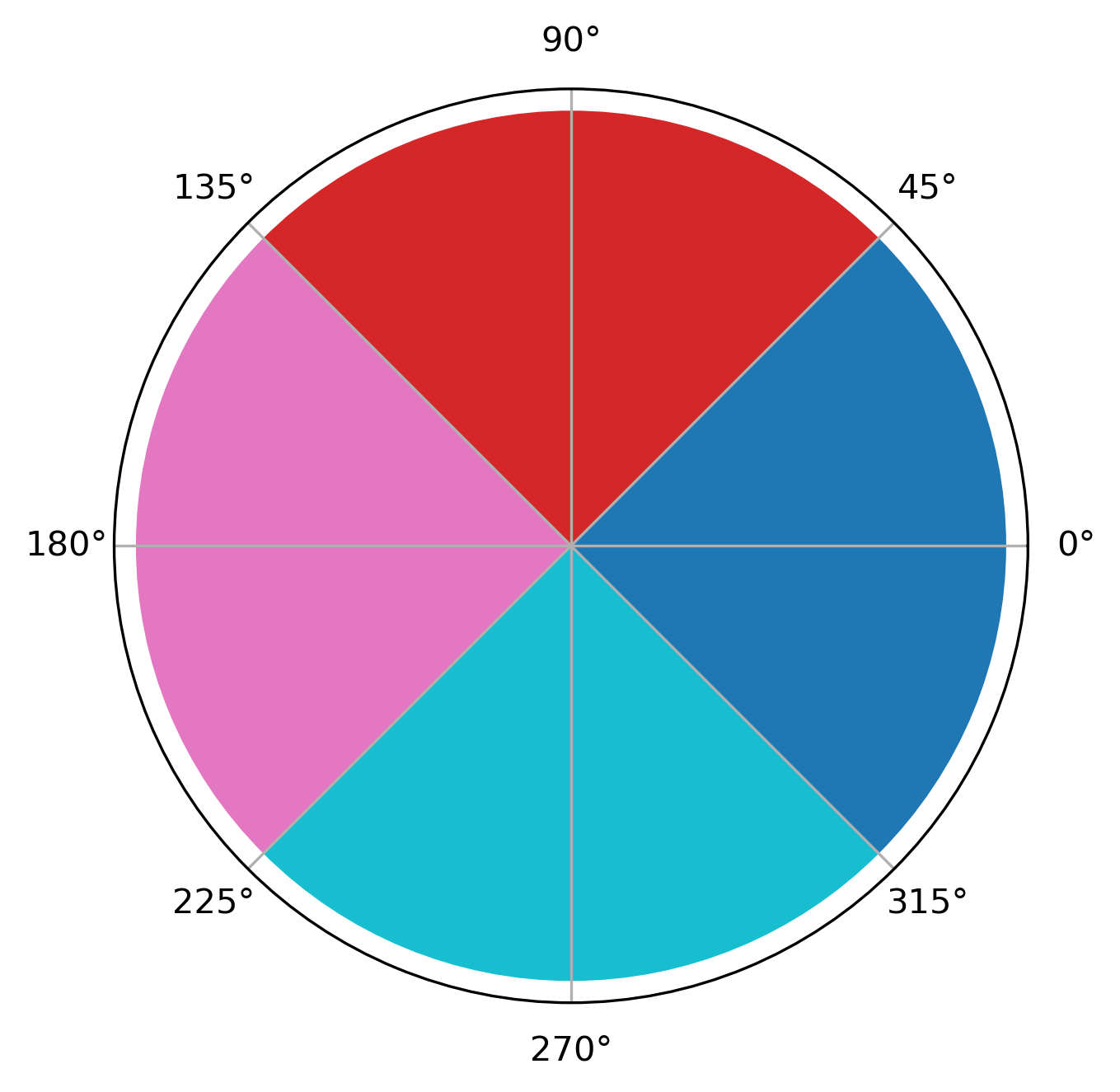}
    \caption{Sections of $\mathbb{C}$ yielded by the $\mathtt{csign}$ function with $K=4$.}
    \label{fig:csign}
\end{figure}

The complex signum activation function, also known as the multivalued threshold function, dates back to the early works of Aizenberg and collaborators on complex-valued neural networks \cite{aizenberg71,aizenberg92}. This function maps a complex number to a root of unity, thus yielding a quantization of the phase. Formally, given a positive integer \( K \), called the \textbf{resolution factor}, the complex signum function is defined by 
\begin{equation}
\label{eq:csign}
\mathtt{csign}_K(z) =
\begin{cases}
1 & 0 \leq \theta < \theta_K, \\
\varepsilon_1 & \theta_K < \theta < 3\theta_K, \\
\vdots & \vdots \\
\varepsilon_{K-1} & (2K-3)\theta_K < \theta < (2K-1)\theta_K, \\
1 & (2K-1)\theta_K < \theta < 2\pi.
\end{cases}
\end{equation}
where $\theta_K = \pi/K$ is know as \textbf{phase quanta}, $\varepsilon_\ell = e^{2\ell\theta_K \ii}$ is the $\ell$th root of the unity, for $\ell=0,\ldots,K-1$, and $\theta$ denotes the phase of $z$ \cite{Aizenberg2011Complex-ValuedNeurons,kobayashi17e}.
Note that the \( \mathtt{csign}_K \) produces \( K \) discrete complex numbers that are uniformly distributed on the unit circle in the complex plane. Figure \ref{fig:csign} illustrates the sections of the complex plane generated by the \( \mathtt{csign} \) function when \( K=4 \). 

In the literature, researchers have proposed complex-valued ``multistate'' neural associative memories using complex-valued Hopfield neural networks (CvHNNs) with the complex signum function \cite{aizenberg92,Jankowski1996Complex-valuedMemory}. The dynamics of the multistate CvHNN are described by \eqref{eq:update}, where $\psi \equiv \mathtt{csign}_K$. It is important to note that the $\mathtt{csign}_K$ activation function is undefined when $\theta = (2\ell - 1)\pi/K$, which occurs at points equidistant from multiple roots of unity. In this situation, the neuron's state remains unchanged \cite{Castro2018SomeNetworks}.

The dynamics of the CvHNN with the complex signum function were extensively studied in literature and applied to image processing \cite{aizenberg92,Jankowski1996Complex-valuedMemory,Muezzinoglu2003AMemory,tanaka09,Castro2018SomeNetworks}. In particular, we have the following theorem concerning the stability of this network \cite{Castro2018SomeNetworks}:
\begin{Theorem} \label{thm:csign}
    The sequence defined by \eqref{eq:update} with $\psi\equiv \mathtt{csign}_K$ is convergent for any initial state in the asynchronous update mode if the synaptic weights satisfy $W_{ij}=\bar{W}_{ji}$ and $W_{ii} \geq 0$ is a non-negative real number.
\end{Theorem}

\subsection{Real-Imaginary-Type Sign Activation Function} \label{subsec:split-sign}

The real-imaginary-type sign activation function $\sign_{\mathbb{C}}:\mathbb{C} \to \mathcal{S}$, also called split-sign complex-valued function, is obtained by applying the real-valued sign function separately at the real and imaginary parts of its argument \cite{Murthy2004Complex-valuedHypercube,Hirose2012Complex-valuedNetworks}. Formally, we have 
\begin{equation} \label{eq:split-sign}
\sign_{\mathbb{C}}(a+b \ii) = \sign_{\mathbb{R}}(a) + \sign_{\mathbb{R}}(b) \ii,
\end{equation}
where $\sign_{\mathbb{R}}:\mathbb{R} \to \{-1,+1\}$ is the real-valued sign function defined by
\begin{equation}
 \sign_{\mathbb{R}}(x) = \begin{cases}
   +1, & x \geq 0, \\
   -1, & x<0.
 \end{cases}
\end{equation}
The states of the neurons of a CvHNN with a split sign activation function belong to the set 
\begin{equation} \label{eq:split-states}
\mathcal{S} = \lbrace +1+\ii, +1-\ii, -1+\ii, -1-\ii \rbrace.
\end{equation}

The nonlinear dynamics of the CvHNN with the real-imaginary-type sign activation function has also been extensively studied in literature \cite{Murthy2004Complex-valuedHypercube,RamaMurthy2004InfiniteHypercube}. In particular, we have the following theorem whose proof can be found in \cite{Murthy2004Complex-valuedHypercube}:
\begin{Theorem}
\label{thm:split-sign}
Consider the CvHNN described by \eqref{eq:update} with the activation function \(\psi \equiv \\splitsign \). If the synaptic weights satisfy $W_{ij}=\bar{W}_{ji}$ and \(W_{ii} \geq 0\) are non-negative real numbers for all \(i,j = 1, \ldots, N\), then the following results hold for the network based on the update mode:
\begin{enumerate}
    \item In serial mode of operation, the network converges to a stable state from any initial state on the complex hypercube \(\mathcal{S}^N\), where \(\mathcal{S}\) is defined by \eqref{eq:split-states}.
    \item In fully parallel mode of operation, the network converges to either a stable state or a cycle of length two, starting from any initial state on \(\mathcal{S}^N\).
\end{enumerate}    
\end{Theorem}

\section{Complex-Valued Hopfield Neural Networks with Magnitude and Phase Quantization}
\label{sec:magnitude_phase}

\begin{figure}
    \centering
    \includegraphics[width=0.95\linewidth]{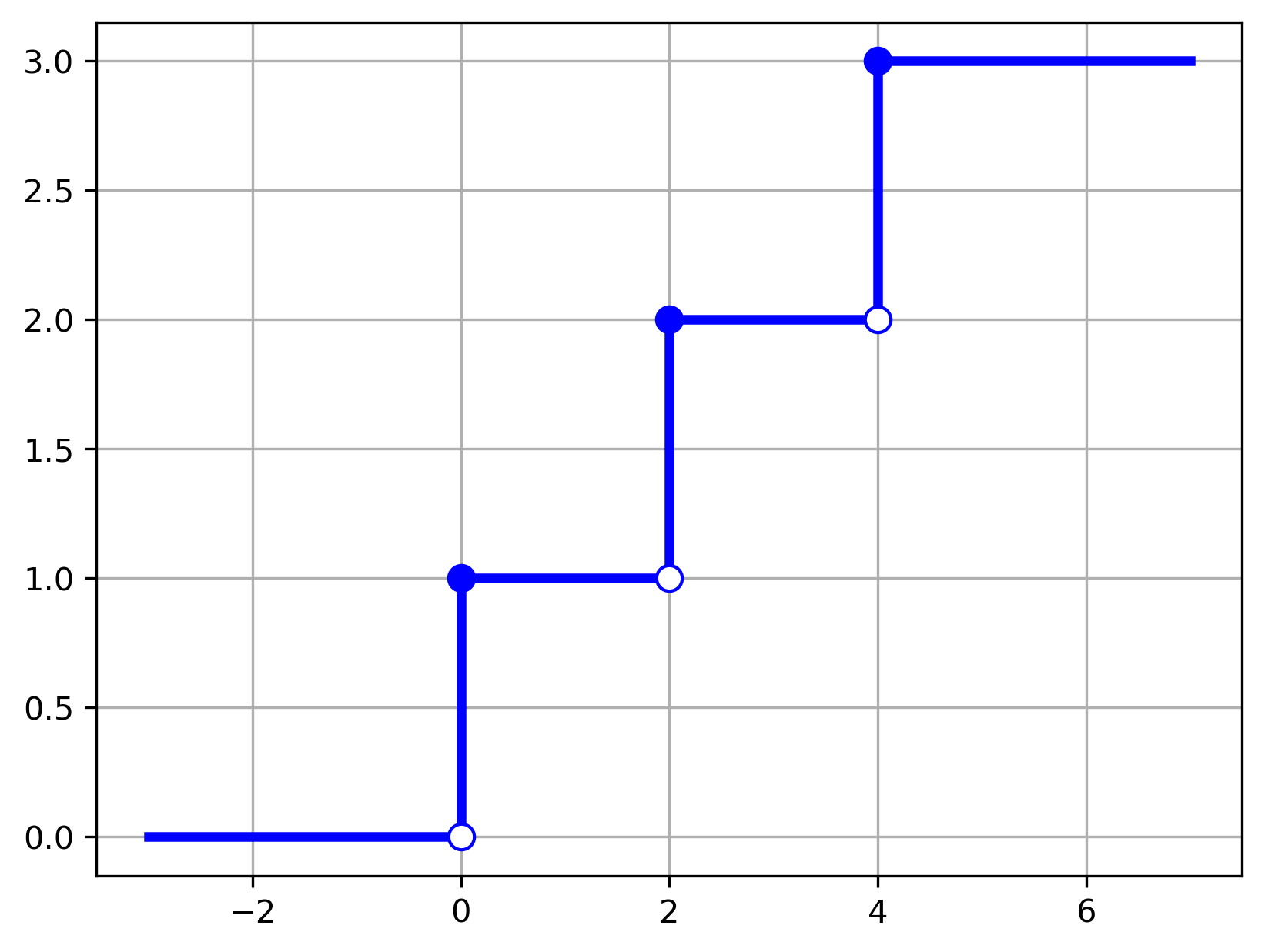}
    \caption{Plot of the $\mathtt{ceil}_{Q,R}$ function with $Q=3$ and $R=2$.}
    \label{fig:ceil}
\end{figure}

Building on previous research on CvHNNs, we propose two new complex-valued neural networks that utilize magnitude and phase quantization. The first network employs Cartesian coordinates, while the second model uses the polar representation of complex numbers. It is important to note that both novel CvHNNs can be used to store multistate patterns. However, they operate in different state spaces, which are defined by the image sets of their activation functions. Specifically, they are also described by \eqref{eq:update} and differ on the activation function $\psi:\mathcal{D} \to \mathcal{S}$, where $\mathcal{D}\subset \mathbb{C}$ is the activation function's domain. The following subsections detail the novel activation functions, which are based on the $\mathtt{ceil}_{Q,R} = \{0,1,\ldots,Q\}$ function defined as follows for fixed positive integers $Q$ and $R$ and any $x \in \mathbb{R}$:
\begin{equation}
    \label{eq:ceil}
    \mathtt{ceil}_{Q,R}(x) =  \begin{cases}
        0, & x < 0, \\
        1, & 0 \leq x < R,\\
        2, & R \leq x < 2R, \\
        \vdots & \vdots \\
        Q, & (Q-1)R \leq x.
    \end{cases}
\end{equation}
Figure \ref{fig:ceil} shows the \(\mathtt{ceil}_{Q,R}\) function with \(Q=3\) and \(R=2\). 

The $\ceil_{Q,R}$ function has been utilized as an activation function in neuron models as noted in \cite{Garimella2019NovelApplications}. Accordingly, the $\ceil_{Q,R}$ function can be written as a superposition of step functions 
\begin{equation}
    \label{eq:step2ceil}
    \ceil_{Q,R}(x) = \sum_{q=1}^Q \step \big(x-(q-1)R\big), \quad \forall x \in \mathbb{R}
\end{equation}
where $\step:\mathbb{R} \to \{0,1\}$ is given by
\begin{equation}
    \label{eq:step-function}
    \step(x) = \begin{cases}
        1, & x \geq 0,\\
        0, & x<0.
    \end{cases}
\end{equation}
As a result, a neuron model incorporating the \(\ceil_{Q,R}\) function can be interpreted as the superposition of \(Q\) linear threshold neurons. These neurons share the same weights but have different bias terms. 

Finally, note that the $\ceil_{Q,R}$ coincides with the $\step$ function when $Q=1$, independently of the value of $R$. Also, note that the \(\ceil_{Q,R}\) function can also be expressed using the \(\sign_{\mathbb{R}}\) function. Specifically, it can be noted that \(\step(x) = (\sign_{\mathbb{R}}(x) + 1)/2\) for any \(x \in \mathbb{R}\).

\subsection{Real-Imaginary-Type Ceiling Activation Function}

The traditional Hopfield neural network is typically described using the $\sign_{\mathbb{R}}$ activation function; however, it was originally conceived with the step function $\step$, as defined in \eqref{eq:step-function}. Additionally, as briefly discussed in \cite{Garimella2019NovelApplications}, a multistate Hopfield-type neural network can be created by replacing the step function with the $\ceil_{Q,R}$ activation function, given in \eqref{eq:ceil}. The dynamics of this resulting real-valued neural network emerge from considering a superposition of traditional Hopfield neural networks that share the same synaptic weights but have different bias terms \cite{Garimella2025OnNeurons}. Inspired by preliminary studies reported in \cite{RamaMurthy2004InfiniteHypercube}, we extend the real-valued Hopfield neural network based on the ceiling neuron to the complex-valued case using a real-imaginary-type activation function.

\begin{figure}
    \centering
    \includegraphics[width=0.95\linewidth]{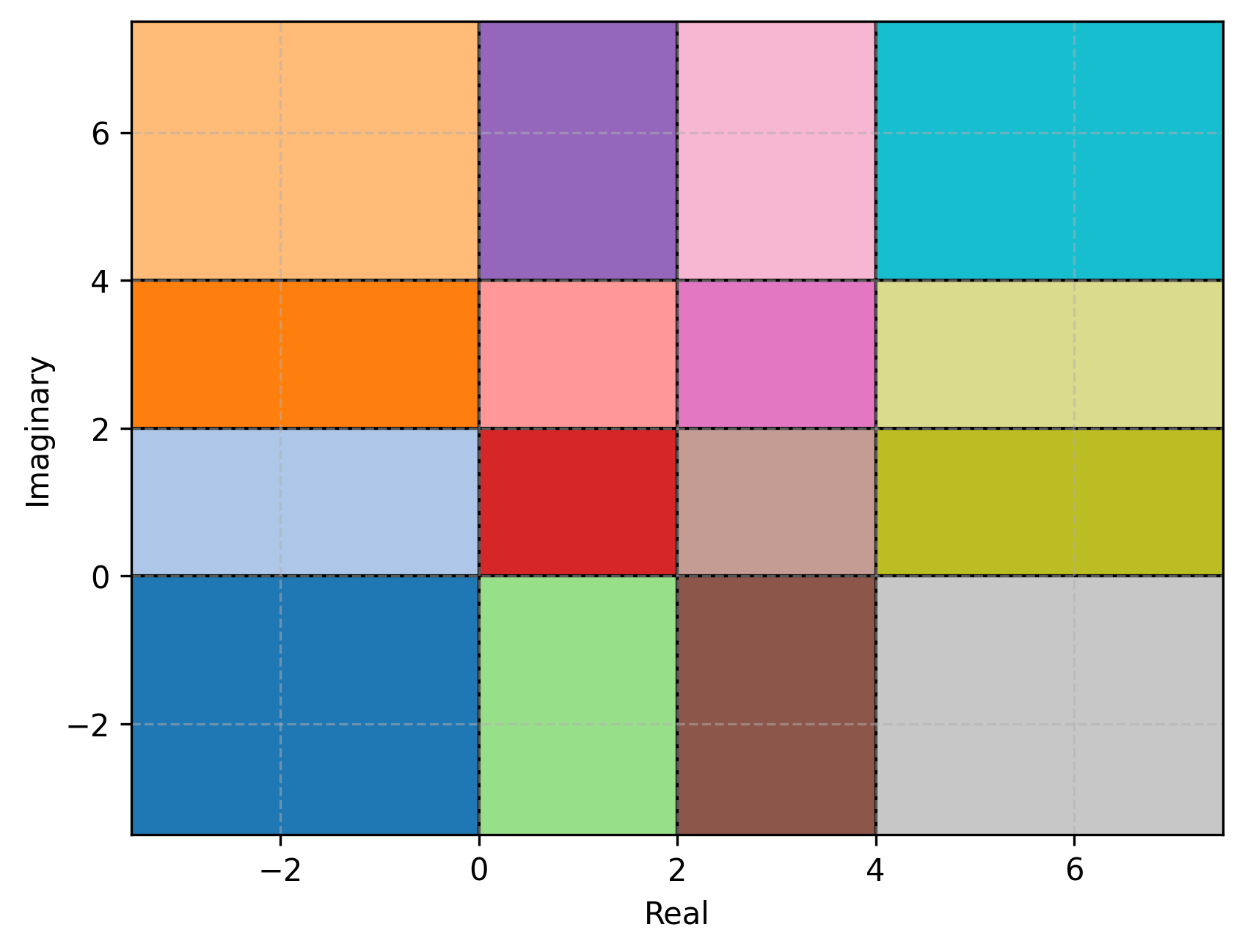}
    \caption{Partition of the complex plane resulting from the $\coceil_{Q,R}$ with $Q=3$ and $R=2$.}
    \label{fig:coceil}
\end{figure}

Precisely, the first CvHNN extends the model described in Section \ref{subsec:split-sign} by replacing the $\sign_{\mathbb{R}}$ function with a ceiling function defined by \eqref{eq:ceil}. The novel model is called CoCeil-CvHNN because its activation function is obtained by applying the ceiling function component-wise (or in a split manner) using the Cartesian representation of a complex number. Formally, the CoCeil-CvHNN is described by \eqref{eq:update} by considering $\psi \equiv \coceil_{Q,R}:\mathbb{C} \to \mathbb{S}_Q$ given by the following equation
\begin{equation}
    \label{eq:CoCeil}
    \coceil_{Q,R}(a + b\ii) = \ceil_{Q,R}(a) + \ceil_{Q,R}(b)\ii,
\end{equation}
where $\mathcal{S}_Q = \{ z = x+ y\ii \in \mathbb{C}: x,y \in \{0,1,\ldots,Q\}\}$ is the image set of the activation function. Figure \ref{fig:coceil} illustrates the partition of the complex plane resulting from the $\coceil_{Q,R}$ with $Q=3$ and $R=2$. Note that the complex plane is divided into $(Q+1)^2$ sections. Therefore, the number of possible states of a CoCeil-CvHNN with $N$ neurons is $(Q+1)^{2N}$, significantly greater than the number of possible states of CvHNN with the real-imaginary-type sign activation function, which is $4^N$.

\subsection{CoSignum Function}

\begin{figure}
    \centering
    \includegraphics[width=0.95\linewidth]{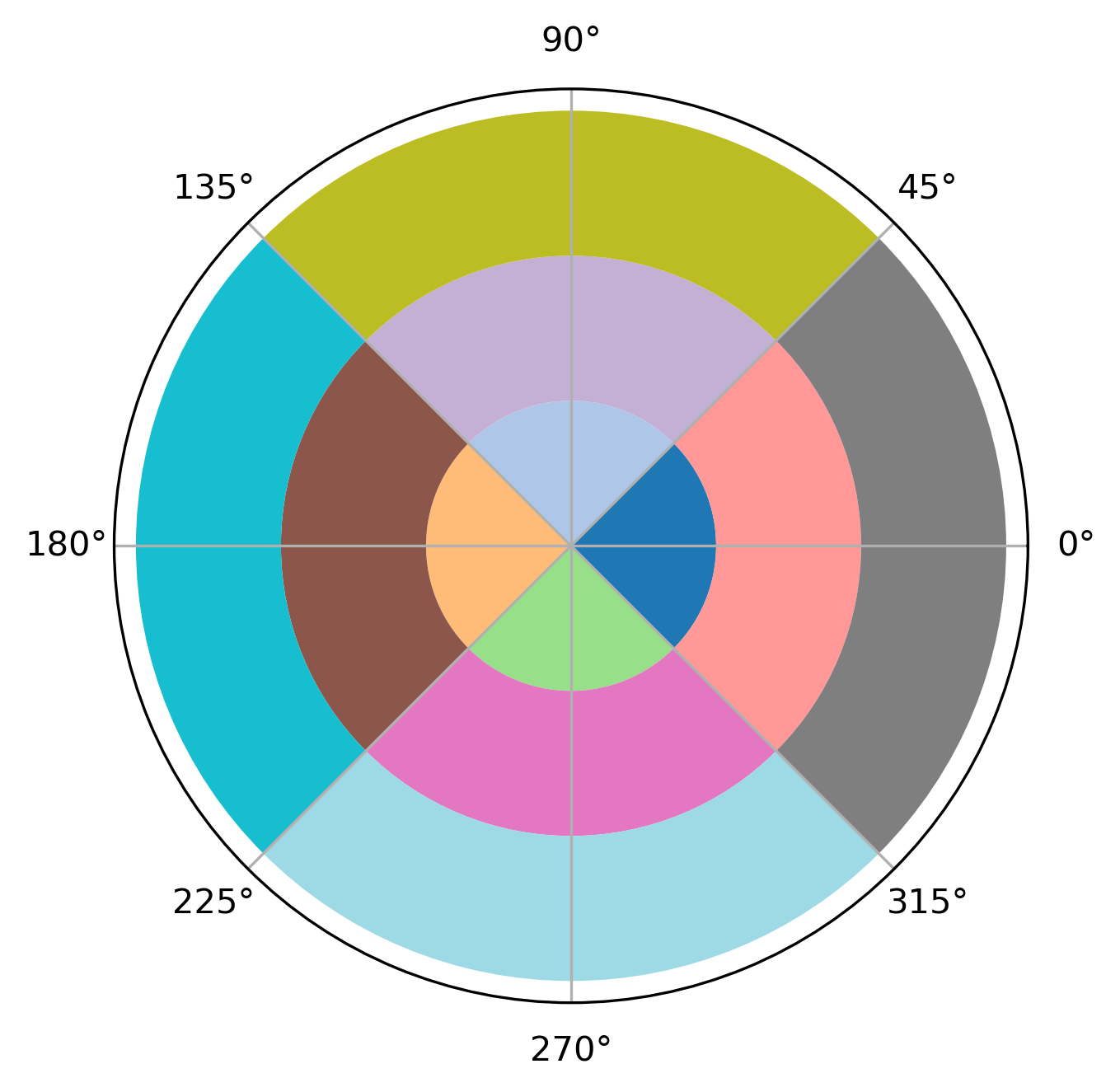}
    \caption{Sections of the complex plane yielded by the $\cosign_{Q,K,R}$ function with $Q=3$, $R=2$, and $K=4$.}
    \label{fig:cosign}
\end{figure}

We can extend the complex signum function $\csign_K$ by incorporating the $\ceil_{Q,R}$ function in the magnitude of a complex number. In other words, the cosignum function, denoted by $\cosign$, extends the complex signum function defined in \eqref{eq:csign} by quantizing both the magnitude and phase. Formally, given positive integers $Q,R,K$, the $\cosign_{Q,R,K}:\mathcal{D} \to \mathcal{S}_{Q,K}$ is defined by
\begin{equation}
    \label{eq:cosign}
    \cosign_{Q,R,K}(z) = \ceil_{Q,R}(|z|)\csign_K(z),
\end{equation}
where the image set of the activation function is
\begin{equation}
\mathcal{S}_{Q,K} = \{ r \varepsilon_\ell: r \in \{1,\ldots,Q\} \text{ and }
\varepsilon_\ell^K = 1\}.    
\end{equation}
Figure \ref{fig:cosign} shows the sections of the complex plane yielded by the $\cosign_{Q,K,R}$ function with $Q=3$, $R=2$, and $K=4$. Note that the complex plane is divided into $QK$ sections. Moreover, the $\cosign_{Q,R,K}$ coincides with $\csign_K$ when $Q=1$, independently of the parameter $R$.

The CoSign-CvHNN, which is based on the $\cosign_{Q,R,K}$ function, is defined by equation \eqref{eq:update}, where $\psi$ is set to $\cosign_{Q,R,K}:\mathcal{D} \to \mathbb{S}_{Q,K}$. Like the complex-signum activation function, the $\cosign_{Q,R,K}$ is undefined when $\theta = (2\ell - 1)\pi/K$, which occurs at points equidistant from multiple roots of unity. In this situation, the neuron's state remains unchanged \cite{castro20nn}. 

Finally, we would like to remark that a CoSign-CvHNN with $N$ neurons admits $(QK)^{N}$ possible states. Hence, the number of possible states for the CoSign-CvHNN is significantly larger than that of the CvHNN with the complex-signum activation function, which has $K^N$ possible states.

\section{Computational Experiments} \label{sec:experiments}

The novel complex-valued activation functions, \( \coceil_{Q,R} \) and \( \cosign_{Q,R,K} \), significantly expand the state space of complex-valued Hopfield neural networks. By employing these two functions, a CvHNN can memorize more stable states than previous models described in the literature while maintaining the same number of neurons. However, for a Hopfield-type network to function as an associative memory, it must always converge to an equilibrium point. Although we do not have proven convergence theorems for the two novel CoCeil- and CoSign-CvHNNs, based on Theorems \ref{thm:csign} and \ref{thm:split-sign}, we conjecture that the synaptic weight matrix must satisfy the conditions \( W_{ij} = \bar{W}_{ji} \) and \( W_{ii} \geq 0 \) for all \( i = 1, \ldots, N \), where \( W_{ii} \) is a non-negative real number. The computational experiments below confirm our conjecture.

First, we generated a synaptic weight matrix whose real and imaginary parts of the entries were randomly selected using a standard normal distribution. Precisely, we sampled $W_{ij}^0 \sim \mathcal{N}(0,1)$ and $W_{ij}^1 \sim \mathcal{N}(0,1)$, where $\mathcal{N}(0,1)$ denotes the standard normal distribution, for all $i,j=1,\ldots,N$. Then, we defined 
\begin{equation}
W_{ij} = \frac{W_{ij}^0+W_{ji}^0}{2} + \frac{W_{ij}^1-W_{ji}^1}{2}\ii \quad \text{and} \quad W_{ii}=0,    
\end{equation}
for all $i,j=1,\ldots,N$. Note that the synaptic weight matrix satisfies the conjectured conditions.

Then, we probed the CoCeil- and CoSign-CvHNN with 5 randomly generated initial states $\boldsymbol{S}(0)=(S_1(0),\ldots,S_N(0))$. Precisely, the CoCeil-CvHNN has been fed by inputs given by
\begin{equation}
    S_i(0) = \coceil_{Q,R}(s_i^0 + s_1^i \ii),
\end{equation}
where $s_i^0,s_i^1 \sim \mathcal{U}(-3,7)$, for all $i=1,\ldots,N$. In other words, $s_i^1$ and $s_i^0$ are samples from a uniform distribution ranging from $-3$ to $7$. Geometrically, $S_i(0)$ is obtained by applying $\coceil_{Q,R}$ on a point sampled uniformly from the square region of the complex plane depicted in Figure \ref{fig:coceil}. Similarly, the initial states of the CoSign-CvHNN are obtained by applying the $\cosign_{Q,R,K}$ function on a point sampled uniformly on the disk shown in Figure \ref{fig:cosign}. Formally, we define 
\begin{equation}
    S_i(0) = \cosign_{Q,R,K}(r_i e^{\theta_i \ii}),
\end{equation}
where $r_i \sim \mathcal{U}(0,R)$ and $\theta_i \in \mathcal{U}(0,2\pi)$, for all $i=1,\ldots,N$.

Finally, we examined the dynamics of the CvHNN by analyzing the energy defined by the following equation:
\begin{equation}
    \label{eq:energy}
    E(\boldsymbol{S}) = -\frac{1}{2} \sum_{i=1}^N \sum_{j=1}^N S_i W_{ij} S_j, \quad \boldsymbol{S} \in \mathcal{S}^N,
\end{equation}
We would like to remark that this energy is also used to prove Theorems \ref{thm:csign} and \ref{thm:split-sign} \cite{Jankowski1996Complex-valuedMemory,Castro2018SomeNetworks,Murthy2004Complex-valuedHypercube}.  

\begin{figure*}
    \centering
    \begin{tabular}{cc}
      a) CoCeil-CvHNN   & b) CoSign-CvHNN \\
          \includegraphics[width=0.48\linewidth]{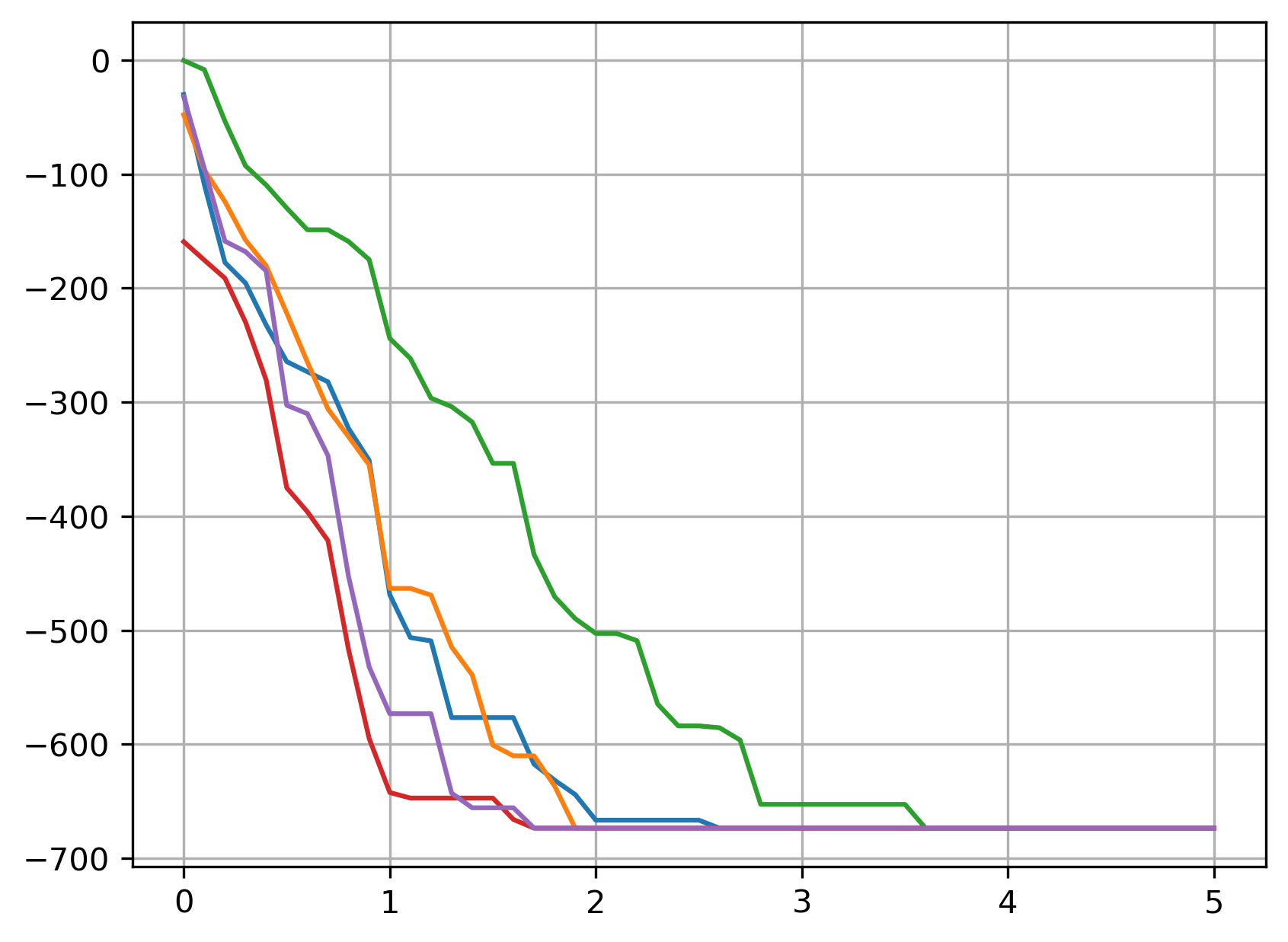}   &
              \includegraphics[width=0.48\linewidth]{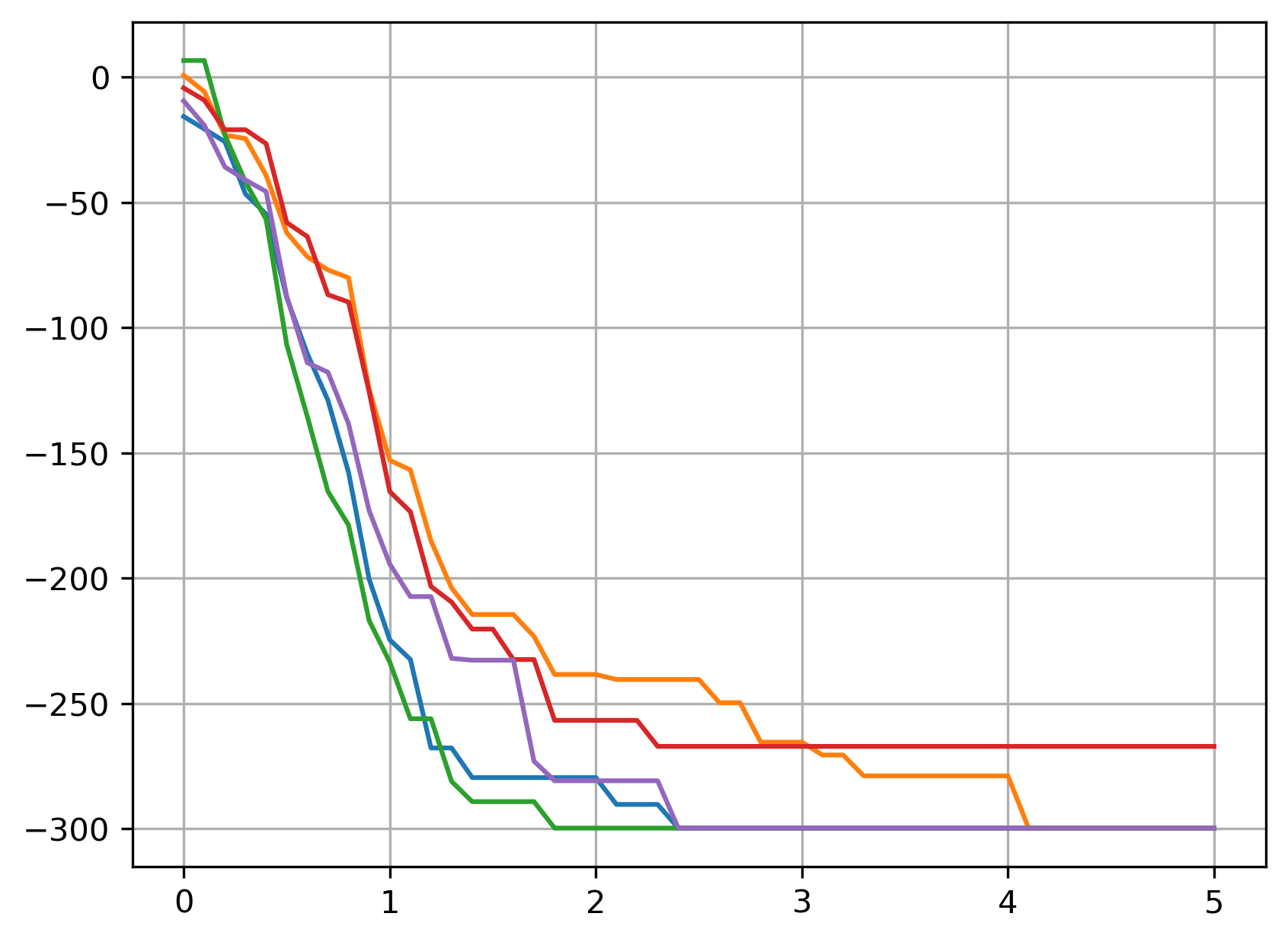}
    \end{tabular}
    \caption{Energy values generated by a) the CoCeil-CvHNN and b) CoSign-CvHNN, starting from five different initial states.}
    \label{fig:Energy}
\end{figure*}

Figure \ref{fig:Energy} illustrates the energy values obtained from the two CvHNN models operating in serial update mode. These plots were generated using parameters \(Q=3\), \(R=2\), \(K=4\), and \(N=10\) neurons. Additionally, each neuron in the network was updated five times, with the x-axis representing the number of updates for each neuron. It is important to note that both the CoCeil-CvHNN and CoSign-CvHNN reached a stationary state when employing synchronous update mode. These results confirm the potential use of the novel CvHNNs as associative memory models.

\section{Concluding Remarks}
\label{sec:concluding}

This paper introduces two novel complex-valued Hopfield neural networks (CvHNNs) that utilize both magnitude and phase quantization to significantly enhance the memory capacity of traditional CvHNN architectures. The first model, referred to as \textit{CoCeil-CvHNN}, employs a ceiling-type activation function in Cartesian coordinates, while the second, \textit{CoSign-CvHNN}, applies a similar function in the polar coordinate system. 
%
By incorporating generalized ceiling functions into the neuron's activation, the proposed networks dramatically expand the state space. Specifically, \textit{CoCeil-CvHNN} achieves a state space of $(Q+1)^{2N}$, and \textit{CoSign-CvHNN} offers $(QK)^N$ states, where $Q$, $R$, and $K$ are quantization parameters and $N$ is the number of neurons. This increase allows for the storage of a greater number of distinct stable patterns, making these networks highly suitable for advanced associative memory applications.

While formal convergence theorems for the new models have not been rigorously proved within the paper, we conjecture stability under specific symmetric weight conditions analogous to those used in the CvHNN described in Section \ref{sec:CvHNNs}. Computational experiments conducted with randomly generated initial states and constrained synaptic weights provide empirical support for this conjecture. Both networks demonstrated convergence to equilibrium points, validating their potential as stable associative memory systems.

Overall, the proposed \textit{CoCeil-} and \textit{CoSign-CvHNNs} offer a promising direction for extending the theoretical and practical capabilities of complex-valued neural networks. Their increased capacity and flexibility make them well-suited for future research and applications involving complex signal processing, pattern recognition, and high-dimensional memory modeling.

\section*{Acknowledgment}
Marcos Eduardo Valle acknowledges financial support from the National Council for Scientific and Technological Development (CNPq), Brazil, under grant no 315820/2021-7, and the São Paulo Research Foundation (FAPESP), Brazil, under grant no 2023/03368-0.



\bibliographystyle{IEEEtran}
\bibliography{references}

\begin{thebibliography}{10}
\providecommand{\url}[1]{#1}
\csname url@samestyle\endcsname
\providecommand{\newblock}{\relax}
\providecommand{\bibinfo}[2]{#2}
\providecommand{\BIBentrySTDinterwordspacing}{\spaceskip=0pt\relax}
\providecommand{\BIBentryALTinterwordstretchfactor}{4}
\providecommand{\BIBentryALTinterwordspacing}{\spaceskip=\fontdimen2\font plus
\BIBentryALTinterwordstretchfactor\fontdimen3\font minus \fontdimen4\font\relax}
\providecommand{\BIBforeignlanguage}[2]{{%
\expandafter\ifx\csname l@#1\endcsname\relax
\typeout{** WARNING: IEEEtran.bst: No hyphenation pattern has been}%
\typeout{** loaded for the language `#1'. Using the pattern for}%
\typeout{** the default language instead.}%
\else
\language=\csname l@#1\endcsname
\fi
#2}}
\providecommand{\BIBdecl}{\relax}
\BIBdecl

\bibitem{mcculloch43}
W.~S. McCulloch and W.~Pitts, ``{A Logical Calculus of the Ideas Immanent in Nervous Activity},'' \emph{Bulletin of Mathematical Biophysics}, vol.~5, pp. 115--133, 1943.

\bibitem{haykin09}
S.~Haykin, \emph{{Neural Networks and Learning Machines}}, 3rd~ed.\hskip 1em plus 0.5em minus 0.4em\relax Upper Saddle River, NJ: Prentice-Hall, 2009.

\bibitem{aizenberg71}
N.~N. Aizenberg, Y.~L. Ivaskiv, and D.~A. Pospelov, ``{A certain generalization of threshold functions},'' \emph{Dokrady Akademii Nauk SSSR}, vol. 196, pp. 1287--1290, 1971.

\bibitem{Aizenberg2011Complex-ValuedNeurons}
I.~Aizenberg, \emph{{Complex-Valued Neural Networks with Multi-Valued Neurons}}, ser. Studies in Computational Intelligence.\hskip 1em plus 0.5em minus 0.4em\relax Berlin, Heidelberg: Springer Berlin Heidelberg, 2011, vol. 353.

\bibitem{Hirose2009Complex-valuedOrigins}
A.~Hirose, ``{Complex-valued neural networks: The merits and their origins},'' \emph{Proceedings of the International Joint Conference on Neural Networks}, pp. 1237--1244, 2009.

\bibitem{Hirose2012Complex-valuedNetworks}
------, ``{Complex-valued neural networks},'' \emph{Studies in Computational Intelligence}, vol. 400, pp. 1--214, 2012.

\bibitem{Trabelsi17complex}
C.~Trabelsi, O.~Bilaniuk, Y.~Zhang, D.~Serdyuk, S.~Subramanian, J.~F. Santos, S.~Mehri, N.~Rostamzadeh, Y.~Bengio, and C.~J. Pal, ``{Deep complex networks},'' 5 2017.

\bibitem{Voigtlaender2023TheNetworks}
F.~Voigtlaender, ``{The universal approximation theorem for complex-valued neural networks},'' \emph{Applied and Computational Harmonic Analysis}, vol.~64, pp. 33--61, 5 2023.

\bibitem{Scardapane2020Complex-ValuedFunctions}
S.~Scardapane, S.~Van~Vaerenbergh, A.~Hussain, and A.~Uncini, ``{Complex-Valued Neural Networks with Nonparametric Activation Functions},'' \emph{IEEE Transactions on Emerging Topics in Computational Intelligence}, vol.~4, no.~2, pp. 140--150, 4 2020.

\bibitem{Lee2022Complex-ValuedSurvey}
C.~Lee, H.~Hasegawa, and S.~Gao, ``{Complex-Valued Neural Networks: A Comprehensive Survey},'' \emph{IEEE/CAA Journal of Automatica Sinica}, vol.~9, no.~8, pp. 1406--1426, 8 2022.

\bibitem{Wolff2025CVKAN:Networks}
\BIBentryALTinterwordspacing
M.~Wolff, F.~Eilers, and X.~Jiang, ``{CVKAN: Complex-Valued Kolmogorov-Arnold Networks},'' 2025. [Online]. Available: \url{https://arxiv.org/abs/2502.02417}
\BIBentrySTDinterwordspacing

\bibitem{Hopfield82}
J.~J. Hopfield, ``{Neural networks and physical systems with emergent collective computational abilities},'' \emph{Proceedings of the National Academy of Sciences of the United States of America}, vol.~79, no.~8, pp. 2554--2558, 1982.

\bibitem{mceliece87}
R.~J. McEliece, E.~C. Posner, E.~R. Rodemich, and S.~S. Venkatesh, ``The capacity of the {Hopfield} associative memory,'' \emph{IEEE Transactions on Information Theory}, vol.~1, pp. 33--45, 1987.

\bibitem{hassoun95}
M.~H. Hassoun, \emph{{Fundamentals of Artificial Neural Networks}}.\hskip 1em plus 0.5em minus 0.4em\relax Cambridge, MA: MIT Press, 1995.

\bibitem{hassoun97}
M.~H. Hassoun and P.~B. Watta, ``{Associative Memory Networks},'' in \emph{Handbook of Neural Computation}, E.~Fiesler and R.~Beale, Eds.\hskip 1em plus 0.5em minus 0.4em\relax Oxford University Press, 1997, pp. C1.3:1--C1.3:14.

\bibitem{Goles-Chacc1985DecreasingNetworks}
E.~Goles-Chacc, F.~Fogelman-Soulie, and D.~Pellegrin, ``{Decreasing energy functions as a tool for studying threshold networks},'' \emph{Discrete Applied Mathematics}, vol.~12, no.~3, pp. 261--277, 11 1985.

\bibitem{shen97}
D.~Shen and H.~H.~S. Ip, ``{A Hopfield neural network for adaptive image segmentation: An active surface paradigm},'' \emph{Pattern Recognition Letters}, vol.~18, no.~1, pp. 37--48, 1997.

\bibitem{smith98}
K.~Smith, M.~Palaniswami, and M.~Krishnamoorthy, ``{Neural Techniques for Combinatorial Optimization with Applications},'' \emph{IEEE Transactions on Neural Networks}, vol.~9, no.~6, pp. 1301--1318, 1998.

\bibitem{pajares06}
G.~Pajares, ``{A Hopfield Neural Network for Image Change Detection},'' \emph{IEEE Transaction on Neural Networks}, vol.~17, no.~5, pp. 1250--1264, 2006.

\bibitem{pajares10}
G.~Pajares, M.~Guijarro, and A.~Ribeiro, ``{A Hopfield Neural Network for Combining Classifiers Applied to Textured Images},'' \emph{Neural Networks}, vol.~23, no.~1, pp. 144--153, 2010.

\bibitem{Tolmachev2020NewMinimisation}
P.~Tolmachev and J.~H. Manton, ``{New Insights on Learning Rules for Hopfield Networks: Memory and Objective Function Minimisation},'' in \emph{Proceedings of the International Joint Conference on Neural Networks}.\hskip 1em plus 0.5em minus 0.4em\relax Institute of Electrical and Electronics Engineers Inc., 7 2020.

\bibitem{Krotov2020LargeLearning}
\BIBentryALTinterwordspacing
D.~Krotov and J.~Hopfield, ``{Large Associative Memory Problem in Neurobiology and Machine Learning},'' 8 2020. [Online]. Available: \url{http://arxiv.org/abs/2008.06996}
\BIBentrySTDinterwordspacing

\bibitem{Ramsauer2020HopfieldNeed}
\BIBentryALTinterwordspacing
H.~Ramsauer, B.~Sch{\"{a}}fl, J.~Lehner, P.~Seidl, M.~Widrich, L.~Gruber, M.~Holzleitner, M.~Pavlovi{\'{c}}, G.~K. Sandve, V.~Greiff, D.~Kreil, M.~Kopp, G.~Klambauer, J.~Brandstetter, and S.~Hochreiter, ``{Hopfield Networks is All You Need},'' 7 2020. [Online]. Available: \url{http://arxiv.org/abs/2008.02217}
\BIBentrySTDinterwordspacing

\bibitem{Karakida2024HierarchicalBreaking}
R.~Karakida, T.~Ota, and M.~Taki, ``{Hierarchical Associative Memory, Parallelized MLP-Mixer, and Symmetry Breaking},'' 6 2024.

\bibitem{noest88a}
A.~J. Noest, ``{Associative Memory in Sparse Phasor Neural Networks},'' \emph{EPL (Europhysics Letters)}, vol.~6, no.~5, p. 469, 1988.

\bibitem{noest88b}
------, ``{Discrete-state phasor neural networks},'' \emph{Physical Review A}, vol.~38, no.~4, pp. 2196--2199, 8 1988.

\bibitem{noest88c}
------, ``{Phasor Neural Networks},'' in \emph{Neural Information Processing Systems}, D.~Z. Anderson, Ed.\hskip 1em plus 0.5em minus 0.4em\relax American Institute of Physics, 1988, pp. 584--591.

\bibitem{Jankowski1996Complex-valuedMemory}
S.~Jankowski, A.~Lozowski, and J.~M. Zurada, ``{Complex-valued multistate neural associative memory},'' \emph{IEEE Transactions on Neural Networks}, vol.~7, no.~6, pp. 1491--1496, 1996.

\bibitem{Muezzinoglu2003AMemory}
M.~K. M{\"{u}}ezzinoǧlu, C.~G{\"{u}}zeli{\c{s}}, and J.~M. Zurada, ``{A new design method for the complex-valued multistate hopfield associative memory},'' \emph{IEEE Transactions on Neural Networks}, vol.~14, no.~4, pp. 891--899, 7 2003.

\bibitem{lee06}
D.-L. Lee, ``{Improvements of complex-valued Hopfield associative memory by using generalized projection rules},'' \emph{IEEE Transactions on Neural Networks}, vol.~17, no.~5, pp. 1341--1347, 2006.

\bibitem{tanaka09}
G.~Tanaka and K.~Aihara, ``{Complex-Valued Multistate Associative Memory With Nonlinear Multilevel Functions for Gray-Level Image Reconstruction},'' \emph{IEEE Transactions on Neural Networks}, vol.~20, no.~9, pp. 1463--1473, 9 2009.

\bibitem{Hashimoto2025EnhancingParallelization}
T.~Hashimoto, T.~Isokawa, M.~Kobayashi, and N.~Kamiura, ``{Enhancing computational efficiency of gradient descent in complex-valued Hopfield neural network through GPU parallelization},'' \emph{Nonlinear Theory and Its Applications, IEICE}, vol.~16, no.~1, pp. 197--207, 2025.

\bibitem{Kobayashi2025Complex-valuedConnections}
M.~Kobayashi, ``{Complex-valued Hopfield associative memories with hybrid connections},'' \emph{Nonlinear Theory and Its Applications, IEICE}, vol.~16, no.~1, pp. 13--29, 2025.

\bibitem{Isokawa2013QuaternionicMemories}
T.~Isokawa, H.~Nishimura, and N.~Matsui, ``{Quaternionic Neural Networks for Associative Memories},'' in \emph{Complex-Valued Neural Networks: Advances and Applications}.\hskip 1em plus 0.5em minus 0.4em\relax John Wiley and Sons, 5 2013, pp. 103--131.

\bibitem{Kobayashi2020SplitNetworks}
M.~Kobayashi, ``{Split Quaternion-Valued Twin-Multistate Hopfield Neural Networks},'' \emph{Advances in Applied Clifford Algebras}, vol.~30, no.~3, pp. 1--9, 7 2020.

\bibitem{Kobayashi2020HopfieldFour-group}
------, ``{Hopfield neural networks using Klein four-group},'' \emph{Neurocomputing}, vol. 387, pp. 123--128, 4 2020.

\bibitem{vallejo08}
J.~R. Vallejo and E.~Bayro-Corrochano, ``{Clifford Hopfield Neural Networks},'' in \emph{2008 IEEE International Joint Conference on Neural Networks (IEEE World Congress on Computational Intelligence)}, 2008, pp. 3609--3612.

\bibitem{castro20nn}
F.~Z. de~Castro and M.~E. Valle, ``{A broad class of discrete-time hypercomplex-valued Hopfield neural networks},'' \emph{Neural Networks}, vol. 122, pp. 54--67, 2020.

\bibitem{Garimella2023Vector-ValuedMemories}
R.~M. Garimella, M.~E. Valle, G.~Vieira, A.~Rayala, and D.~Munugoti, ``{Vector-Valued Hopfield Neural Networks and Distributed Synapse Based Convolutional and Linear Time-Variant Associative Memories},'' \emph{Neural Processing Letters}, vol.~55, no.~4, pp. 4163--4182, 8 2023.

\bibitem{Murthy2004Complex-valuedHypercube}
G.~R. Murthy and D.~Praveen, ``{Complex-valued neural associative memory on the complex hypercube},'' in \emph{2004 IEEE Conference on Cybernetics and Intelligent Systems}, 2004, pp. 648--652.

\bibitem{RamaMurthy2004InfiniteHypercube}
G.~Rama~Murthy, ``{Infinite Population, Complex Valued State Neural Network on the Complex Hypercube},'' in \emph{International Conference on Cognitive Science (ICCS 2004)}, 2004.

\bibitem{aizenberg92}
N.~N. Aizenberg and I.~N. Aizenberg, ``{CNN based on multivalued neuron as a model of associative memory for gray-scale images},'' in \emph{Proceedings of the 2nd International Workshop on Cellular Neural Networks and Their Applications}, 1992, pp. 36--42.

\bibitem{Castro2018SomeNetworks}
F.~Z.~d. Castro and M.~E. Valle, ``{Some Remarks on the Stability of Discrete-Time Complex-Valued Multistate Hopfield Neural Networks},'' in \emph{Proceeding Series of the Brazilian Society of Computational and Applied Mathematics}, vol.~6, no.~2.\hskip 1em plus 0.5em minus 0.4em\relax SBMAC, 12 2018.

\bibitem{Bruck1988ANetworks}
J.~Bruck and J.~Goodman, ``{A generalized convergence theorem for neural networks},'' \emph{IEEE Transactions on Information Theory}, vol.~34, no.~5, pp. 1089--1092, 9 1988.

\bibitem{kobayashi17e}
M.~Kobayashi, ``{Symmetric Complex-Valued Hopfield Neural Networks},'' \emph{IEEE Transactions on Neural Networks and Learning Systems}, vol.~28, no.~4, pp. 1011--1015, 2017.

\bibitem{Garimella2019NovelApplications}
R.~M. Garimella, S.~D. Munugoti, and A.~Rayala, ``{Novel Ceiling Neuron Model and its Applications},'' in \emph{2019 International Joint Conference on Neural Networks (IJCNN)}, vol. 2019-July.\hskip 1em plus 0.5em minus 0.4em\relax IEEE, 7 2019, pp. 1--8.

\bibitem{Garimella2025OnNeurons}
R.~M. Garimella and F.~Z. Castro, ``{On the dynamics of Real Valued Hopfield-Type Neural Networks based on Ceiling Neurons},'' 2025, submitted for publication.

\end{thebibliography}

\end{document}